\documentclass[a4 paper, conference]{IEEEtran}

\usepackage[a4paper,top=54pt,left=38pt,right=38pt,bottom=116pt]{geometry}

\usepackage{url}
\usepackage{amsmath,amsfonts,amssymb,amsthm}
\usepackage{graphicx}
\usepackage{epsfig,epstopdf}
\usepackage[lined,ruled,commentsnumbered]{algorithm2e}
\usepackage{color}
\usepackage{balance}

\definecolor{az}{rgb}{0.0,0.0,0.0}
\definecolor{jc}{rgb}{0.0,0.0,0.0}
\definecolor{msb}{rgb}{0.0,0.0,0.0}

\newcommand{\az}{\color{az}}
\newcommand{\jc}{\color{jc}}
\newcommand{\msb}{\color{msb}}

\newcommand{\cm}{.1 cm}

\graphicspath{{./fig/}}

\newtheorem{prop}{Proposition}

\newtheorem{lm}{Lemma}

\title{Kernelized Covariance for Action Recognition}

\author{\IEEEauthorblockN{{\jc Jacopo Cavazza}\IEEEauthorrefmark{1}\IEEEauthorrefmark{2}, {\az Andrea Zunino}\IEEEauthorrefmark{1}\IEEEauthorrefmark{2}, {\msb Marco San Biagio}\IEEEauthorrefmark{1} and Vittorio Murino\IEEEauthorrefmark{1}\IEEEauthorrefmark{3}}
\IEEEauthorblockA{\IEEEauthorrefmark{1} Pattern Analysis \& Computer Vision -- Istituto Italiano di Tecnologia, Via Morego 30, 16163, Genova, Italy}
\IEEEauthorblockA{\IEEEauthorrefmark{2} Universit\`{a} degli Studi di Genova -- Dipartimento di Ingegneria Navale, Elettrica, Elettronica e delle Telecomunicazioni, \\  Via All'Opera Pia, 11A, 16145, Genova, Italy}
\IEEEauthorblockA{\IEEEauthorrefmark{3} Universit\`{a} di Verona -- Dipartimento di Informatica, Strada le Grazie 15, 37134, Verona, Italy}
\IEEEauthorblockN{\texttt{\{jacopo.cavazza,andrea.zunino,marco.sanbiagio,vittorio.murino\}@iit.it}}}

\begin{document}

\maketitle


\begin{abstract} In this paper we aim at increasing the descriptive power of the covariance matrix, limited in capturing linear mutual dependencies between variables only. We present a rigorous and principled mathematical pipeline to recover the kernel trick for computing the covariance matrix, enhancing it to model more complex, non-linear relationships conveyed by the raw data. To this end, we propose \textit{\textbf{Kernelized-COV}}, which generalizes the original covariance representation without compromising the efficiency of the computation. In the experiments, we validate the proposed framework against many previous approaches in the literature, scoring on par or superior with respect to the state of the art on benchmark datasets for 3D action recognition.
\end{abstract}

\IEEEpeerreviewmaketitle

\begin{figure*}[t]
	\vspace{\cm}
	\includegraphics[width=\textwidth,keepaspectratio]{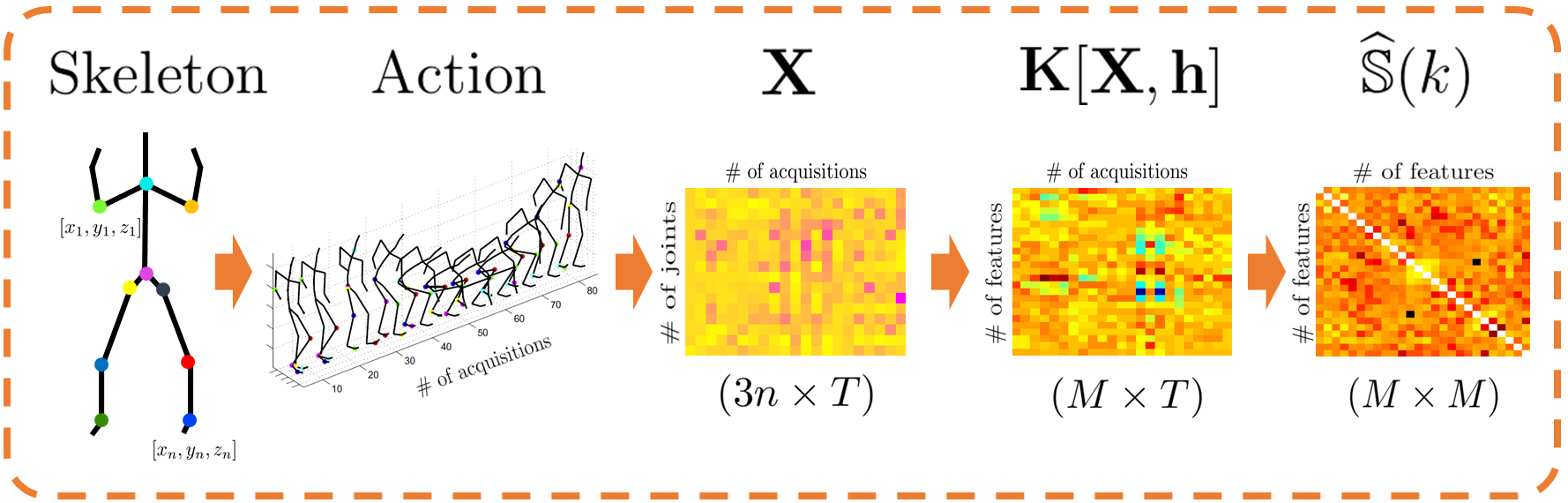}
	\caption{Overview of the proposed framework. From the human skeleton, for each action, we extract MoCap data. The latter are represented through the matrix $\mathbf{X}$ which collects the three-dimensional coordinates, referring to the $n$ joints, acquired during $T$ successive instants. A kernel encoding is performed by means of the Gram matrix $\mathbf{K}[\mathbf{X},\mathbf{h}],$ which is finally used to compute the kernelized covariance $\widehat{\mathbb{S}}(k)$.}
	\label{fig:pipeline}
	\vspace{1 cm}
\end{figure*} 

\vspace{.2 cm}

\noindent {\footnotesize {\bf Publicly available code:} \url{https://www.iit.it/pavis/code/kcar}}


\section{Introduction}\label{sez:intro}

In the past three decades, motion capture systems -- MoCap -- have been engineered with the ultimate goal of tracking and recording human motion while guaranteeing high resolutions in both spatial and temporal domains. The acquired data consist of time series of joint/marker 3D positions and are broadly used for several different applications, \textit{e.g.}, studying human motions in sport sciences, inferring biometric patterns for person identification or generating realistic motion sequences in computer animation to name a few \cite{survey}. Among these ones, action and activity recognition displays a crucial role in human-robot interaction, autonomous driving vehicles and video-surveillance \cite{survey2}. However, devising effective methods to analyze MoCap data is demanding due to the many yet unsolved problems related, for instance, to missing acquisitions {\jc of joints coordinates} or to highly corrupted data.

Previous attempts to face these issues either rely on some distance learning techniques (\textit{e.g.}, subspace view invariant metric \cite{Sheikh:ICCV05}) or applied stochastic techniques to model the degree of uncertainty in the data. For instance, a hidden Markov model is used in \cite{Lv:ECCV06} to produce weak classifiers which are enhanced by AdaBoost. Furthermore, \cite{Li:CVPRw10} proposed an action graph to model the dynamics for action recognition and exploited a bag of 3D points as feature representation.

Since the spatial and/or temporal dimensions of the recorded data can be heavy, dimensionality reduction \cite{Yang:CVPRw12} or feature selection \cite{Ofli:CVPRw13} methods have been devised. However, in general, the classification is subsequent to a design phase of discriminative features such as actionlets \cite{Wang:CVPR12}, random occupancy patterns \cite{Wang:ECCV12}, pose-based sets \cite{pose}, space-time trajectories \cite{DB_B_3D}, velocity and acceleration \cite{movpose}, normal vectors \cite{SNV} or Lie group geometry embedding \cite{Vemulapalli:CVPR14}. 

As a different paradigm to a customized class of task-specific features, generalizable representations driven by covariance matrix were shown to be promising, either encoding spatio-temporal derivatives of joint positions \cite{Sanin:2013} or producing a hierarchical temporal pyramids of descriptors \cite{egizi}. 

Recently, the new state of the art for action and activity recognition from MoCap data was set by \cite{beyond}, where several Gram matrices are computed to produce multiple representations of the joint positions of each trial and, once a fusion step is performed, a log-Euclidean kernel feeds the SVM classifier. Therein, the covariance is replaced by kernel matrices and this is motivated by the observation that the former can only understand linear relationships while the latter allows to model general ones. In this work, we pursue an opposite perspective, focusing on the covariance representation and rigorously devising a \textit{kernelized} version to extend its discriminative power.

Indeed, by the direct usage of a kernel, we can avoid any preliminary explicit feature encoding (as, for instance, occurs in \cite{Sanin:2013}) and, for a general class of kernel functions, we recover the kernel trick for covariance matrix estimation. As a result, its descriptiveness increases from linear to arbitrary relationships modelling, while the efficiency in the computation is preserved. 

To the best of our knowledge, this problem was never faced before in this principled way in both machine learning and pattern recognition fields.

To sum up, we highlight the contributions of this paper.
\begin{itemize}
	\item We propose a new kernelized representation for covariance matrix, namely \textbf{\textit{Kernelized-COV}}. By recovering the well-known kernel trick, we can capture more general interdependencies between variables in a way that the usual covariance descriptor becomes a particular case and the overall computational cost does not increase.
	\item In order to prove the effectiveness of our approach for action and activity recognition of MoCap data, we compare our method against different ones on MSR-Action3D \cite{Action3D}, MSR-Daily-Activity \cite{Daily}, MSRC-Kinect12 \cite{MSRC} and HDM-05 \cite{HDM-05} benchmark datasets. With respect to the state-of-the-art methods \cite{beyond}, the registered performance shows comparable results in the first two datasets and better scores in the remaining ones. This properly certifies that our kernelization is able to bridge the gap between covariance and kernel-based representation.
\end{itemize} 

The rest of the paper is outlined as follows. In Section \ref{sez:back}, we sketch some theoretical background about the covariance matrix. In Section \ref{sez:method}, we present our framework which is experimentally validated in Section \ref{sez:exp}. Finally, Section \ref{sez:conc} draws some conclusions and profiles future work.

\section{Background}\label{sez:back}

At an arbitrary timestamp $t$, a generic MoCap system represents the body of a human agent as the collection $\mathbf{x}(t) \in \mathbb{R}^{3n}$ of the three-dimensional locations $\mathbf{x}_1(t),\dots,\mathbf{x}_n(t)$ of $n$ joints/markers positions, being $\mathbf{x}_i(t) = [x_i(t),y_i(t),z_i(t)]^\top \in \mathbb{R}^3$ the $x,y$ and $z$ coordinates for $i = 1,\dots,n.$ In order to quantify how much any pair of the coordinates mutually change in time, the notion of covariance is classically exploited in statistics \cite{Hamilton}. However, it cannot be computed in absence of a known distribution for the probability according to which the samples $\mathbf{x}(t)$ are drawn. However, this assumption is seldom verified in real cases and, as an alternative, the sampling covariance matrix $\widehat{\mathbb{S}}$ is usually exploited: this is due to the fact that it is an unbiased estimator of the original covariance\footnote{For convenience, in the following, we will concisely refer to the estimator $\widehat{\mathbb{S}}$ as the covariance itself, omitting the ``sampling'' attribute.} and can be computed using a finite number of samples $\mathbf{x}(t)$, $t = 1,\dots,T$, only. Precisely, it is defined as
\vspace{\cm}
\begin{equation}
\label{eq:sampCOV}
\widehat{\mathbb{S}}(\mathbf{X}) = \dfrac{1}{T-1} \sum_{t=1}^T (\mathbf{x}(t) - \boldsymbol{\mu})(\mathbf{x}(t) - \boldsymbol{\mu})^\top,
\vspace{\cm}
\end{equation}
where $\mathbf{X}$ represents the $3n \times T$ data matrix which stacks by columns all the temporal acquisitions $\mathbf{x}(1),\dots,\mathbf{x}(T),$ whose average is denoted by $\boldsymbol{\mu}.$ In matrix notation, \eqref{eq:sampCOV} becomes\footnote{\jc For a matter of space, the technical proof of deriving equation \eqref{eq:sampCOVmat} from \eqref{eq:sampCOV} was moved to the Supplementary Material.}

\vspace{\cm}
\begin{equation}
\label{eq:sampCOVmat}
\widehat{\mathbb{S}}(\mathbf{X}) = \mathbf{XPX}^\top,
\vspace{\cm}
\end{equation}
once defined $\mathbf{P}$ as the $T \times T$ matrix whose $(s,t)$-th entry is
\vspace{\cm}
\begin{equation}\label{eq:P}
\mathbf{P}_{ss} = \frac{1}{T} \qquad \mbox{and} \qquad \mathbf{P}_{st} = -\frac{1}{T^2-T} \quad \mbox{if} \; s \neq t.
\vspace{\cm}
\end{equation}

The usage of the covariance $\widehat{\mathbb{S}}$ to produce descriptors for classification tasks has been intensively studied \cite{TPM:ECCV06,Gabor,Tosato:2013,hasc,MsB,Minh,Harandi:CVPR14,beyond}. In particular, \cite{TPM:ECCV06} proposed patch-specific covariance descriptors, efficiently computed with integral images. Other approaches rely on covariance to systematically encode mutual relationships inside the data and such idea was applied to many different applications such as face recognition \cite{Gabor}, person identification \cite{Tosato:2013} and more general classification tasks \cite{hasc}. Further, covariance was proposed to measure similarities across data samples \cite{MsB}.

This latter direction actually grounds on the mathematical properties of positive definite matrices, exploiting Riemannian metrics on manifold  for image classification: once moved from a finite to an infinite dimensional space, the performance enhances \cite{Minh,Harandi:CVPR14} and only recently deep learning approaches have shown to be superior. However, one of the main limitation related to covariance matrix is that it only enables to capture linear inter-relationships \cite{Hamilton}. For instance, principal component analysis actually exploits a covariance matrix to remove linear correlation of data points \cite{bishop}. Among the attempts for modeling more complicated relationships, additional statistics, such as entropy and mutual information \cite{hasc}, and kernels \cite{beyond} have been adopted. As a different paradigm, one can model non-linear behaviors by preliminary applying a preprocessing step and encode raw data by means of a transformation which increases the feature space. For instance, \cite{Sanin:2013} applied such idea for spatial and temporal derivatives for gesture recognition, \cite{hasc} considered both different color spaces and edge detectors for image classification, and \cite{Tosato:2013} used filter bank responses as features to estimate head orientation. In this latter approach, once defined the feature map $\Phi$ and the transformed data matrix $\boldsymbol{\Phi}(\mathbf{X})$ whose $t$-th column is $\Phi(\mathbf{x}(t))$, the covariance \eqref{eq:sampCOVmat} is now expressed by

\vspace{\cm}
\begin{equation}\label{eq:sampCOVmatPhi}
\widehat{\mathbb{S}}(\boldsymbol{\Phi}(\mathbf{X})) = \boldsymbol{\Phi}(\mathbf{X})\mathbf{P}\boldsymbol{\Phi}(\mathbf{X})^\top.
\vspace{\cm}
\end{equation}
Despite $\widehat{\mathbb{S}}(\boldsymbol{\Phi}(\mathbf{X}))$ is able to capture general relationships embedded in the raw data $\mathbf{X},$ the main bottleneck with \eqref{eq:sampCOVmatPhi} is the requirement of explicit computation for $\boldsymbol{\Phi}(\mathbf{X})$. Indeed, due to feature space augmentation performed by $\Phi,$ the higher dimensionality of such a matrix is more demanding in terms of both storage and computational cost required to calculate \eqref{eq:sampCOVmatPhi} instead of \eqref{eq:sampCOVmat}. Additionally, although infinite feature spaces are common for many classes of feature maps (\textit{e.g.}, the one corresponding to a Gaussian kernel), this case has to be excluded in \eqref{eq:sampCOVmatPhi} since $\boldsymbol{\Phi}(\mathbf{X})$ is infinite dimensional and therefore impossible to compute exactly. In the following Section, we will face the problem of obtaining $\widehat{\mathbb{S}}$ without involving $\boldsymbol{\Phi}(\mathbf{X})$.

\vspace{\cm}
\section{Method}\label{sez:method}

Leveraging on the theory of kernel methods \cite{Sc:2002}, every symmetric and positive definite kernel function $k \colon \mathbb{R}^{3n} \times \mathbb{R}^{3n} \to \mathbb{R}$ can be expressed as

\vspace{\cm} 
\begin{equation}\label{eq:dot}
k(\mathbf{x}, \mathbf{z}) = \langle \Phi(\mathbf{x}), \Phi(\mathbf{z}) \rangle_{\mathcal{H}},
\vspace{\cm}
\end{equation}
where the inner product is computed in the Hilbert space\footnote{For additional details about $\mathcal{H}$ as well as for an extended presentation of the proposed method, please, refer to the Supplementary Material.} $\mathcal{H}$ which defines the range of the feature map $\Phi \colon \mathbb{R}^{3n} \to \mathcal{H}$.
In \eqref{eq:dot}, the kernel trick \cite{Sc:2002} replaces the arbitrary relationships in the original data space with a linear reformulation in $\mathcal{H}$: most importantly, $\Phi$ can be actually skipped, since only requiring the computation of the kernel $k$ (\textit{e.g.}, this happens for support vector machines \cite{bishop}). In our case, we will employ $k$ to obtain the representation $\widehat{\mathbb{S}}(k)$, equivalent to \eqref{eq:sampCOVmatPhi}, that is $\widehat{\mathbb{S}}(k) = \widehat{\mathbb{S}}(\Phi(\mathbf{X}))$, while also skipping the computation of $\Phi$. The following statement moves the first step in this direction.

\vspace{\cm}
\begin{lm}
	\label{th:1}
	Assume that there exist $\mathbf{h}_j \in \mathbb{R}^{3n}$ such that $\Phi(\mathbf{h}_j) = \mathbf{e}_j$ for every $j = 1,\dots,\dim(\mathcal{H})$, being $\mathbf{e}_j$ the unitary element of the canonical base of $\mathcal{H}$ as a vectorial space. Then, there exists a $\dim(\mathcal{H}) \times T$ matrix $\mathbf{K}[\mathbf{X},\mathbf{h}],$ depending only on the kernel $k$, the data $\mathbf{X}$ and $\mathbf{h}_j$, such that, if we define $\widehat{\mathbb{S}}(k) = \mathbf{K}[\mathbf{X},\mathbf{h}]\mathbf{P}{\mathbf{K}[\mathbf{X},\mathbf{h}]}^\top,$ we get $\widehat{\mathbb{S}}(k) = \widehat{\mathbb{S}}(\mathbf{\Phi(X}))$.
\end{lm}

\proof Using \eqref{eq:sampCOVmatPhi}, the $(i,j)$-th entry of $\widehat{\mathbb{S}}(\mathbf{\Phi(X)})$ rewrites
\vspace{\cm}
\begin{equation}\label{eq:tutu}
\widehat{\mathbb{S}}_{ij}(\mathbf{\Phi(X)}) \hspace{-.07cm} = \hspace{-.07cm} \hspace{-.07cm}\sum_{s,t=1}^T {\langle \Phi(\mathbf{x}(s)), \mathbf{e}_i \rangle}_{\mathcal{H}} {\mathbf{P}}_{st} \langle {\Phi(\mathbf{x}(t)), \mathbf{e}_j \rangle}_{\mathcal{H}}.
\vspace{\cm}
\end{equation}
In \eqref{eq:tutu}, once exploited the assumption that $\Phi(\mathbf{h}_j) = \mathbf{e}_j,$ for some $\mathbf{h}_j,$ we can define the $\dim(\mathcal{H}) \times T$ matrix $\mathbf{K}[\mathbf{X},\mathbf{h}]$ whose $(i,s)$-th entry $k(\mathbf{x}(s),\mathbf{h}_i)$ is $\langle \Phi(\mathbf{x}(s)), \mathbf{e}_i \rangle_{\mathcal{H}} = \langle \Phi(\mathbf{x}(s)), \Phi(\mathbf{h}_i) \rangle_{\mathcal{H}}$ and consequently we deduce

\vspace{\cm}
\begin{equation}\label{eq:Sk}
\widehat{\mathbb{S}}(k) = \mathbf{K}[\mathbf{X},\mathbf{h}]\mathbf{P}{\mathbf{K}[\mathbf{X},\mathbf{h}]}^\top \hspace{-.1 cm}= \widehat{\mathbb{S}}(\mathbf{\Phi(X)}),
\vspace{\cm}
\end{equation}
which proves the thesis.
\endproof

Lemma \ref{th:1} certifies that we are able to compute the covariance in terms of the sole kernel $k.$ However, some issues pertain to  the practical feasibility of the assumption
\vspace{\cm}
\begin{equation}
\label{eq:assumption}
\Phi(\mathbf{h}_j) = \mathbf{e}_j,  
\end{equation}
for any $j$, which is nevertheless fundamental for our purposes. 

\vspace{\cm}
Actually, \eqref{eq:assumption} is quite restrictive since the range of $\Phi$ is forced to contain the whole canonical base of $\mathcal{H}$. For instance, if $\mathcal{H} = \mathbb{R}^M,$ \eqref{eq:assumption} consists in a set of $M$ equations that have to be solved in an $M$-dimensional space and, even if we assume that $\Phi(\mathbf{x}) = \mathbf{x},$ the resulting linear system can be either undetermined or impossible. Clearly, in case of a more general shape for $\Phi$, it is not trivial to check whether the assumption \eqref{eq:assumption} is verified. Hence, it seems natural to opt for a different feature map, which can replace $\Phi$ in generating the kernel function $k$, also satisfying \eqref{eq:assumption}. Thus, in the rest of the paper, we will focus on a specific class of stochastic feature maps $\boldsymbol{\Psi}$, actually fulfilling hypothesis \eqref{eq:assumption}, so that the induced linear kernel approximates $k$ in a both stochastic and analytical sense. Therefore, we select the family of functions
\vspace{\cm}
\begin{equation}
\label{eq:serieker}
k(\mathbf{x},\mathbf{z}) = \sum_{\ell = 0}^\infty a_\ell {\langle \mathbf{x}, \mathbf{z} \rangle}^\ell
\vspace{\cm}
\end{equation}
where the dot product $\langle \mathbf{x}, \mathbf{z} \rangle$ is computed in $\mathbb{R}^{3n}$ and $a_\ell \geq 0$ for any $\ell.$ It is worth nothing that,  due to the non-negativeness of these coefficients, since a linear combination of kernels is still positive definite, then \eqref{eq:serieker} admits the representation \eqref{eq:dot}. Also, \eqref{eq:serieker} covers both finite and infinite linear combinations and therefore is comprehensive of a broad class of kernel functions. For instance, it is easily checked that \eqref{eq:serieker} generalizes both the polynomial kernel $k(\mathbf{x},\mathbf{z}) = {\langle \mathbf{x},\mathbf{z} \rangle}^\ell + a_0$ and the exponential-dot product kernel $k(\mathbf{x},\mathbf{z}) = \exp \left( \dfrac{\langle \mathbf{x},\mathbf{z} \rangle}{\sigma^2} \right)$, $\sigma > 0$. In this setting, we now introduce the following lemma which gives the fundamental tool to construct $\boldsymbol{\Psi}$. 

\vspace{\cm}
\begin{lm}
	\label{lm:lm}
	Let $\boldsymbol{\omega} = [\omega_1,\dots,\omega_{3n}]$ a collection of $3n$ independent samples jointly distributed as a mixture of discrete Dirac's deltas and define $\psi(\mathbf{x}) = \langle\boldsymbol{\omega},\mathbf{x}\rangle.$ Then, the expectation of $\psi(\mathbf{x})\psi(\mathbf{z})$ under the distribution of $\boldsymbol{\omega}$ is 
	\vspace{\cm}
	\begin{equation}
	\mathbb{E}_{\boldsymbol{\omega}}[\psi(\mathbf{x})\psi(\mathbf{z})] = \langle \mathbf{x},\mathbf{z} \rangle.
	\vspace{\cm}
	\end{equation}
\end{lm}

\proof
Using the definition of $\psi,$ the property of the mixture of Dirac's delta distribution and the linearity of the expectation $\mathbb{E}_{\boldsymbol{\omega}},$ the thesis comes after the following chain of equivalences

\vspace{\cm}
\begin{align}
\mathbb{E}_{\boldsymbol{\omega}}[\psi(\mathbf{x})\psi(\mathbf{z})] = \mathbb{E}_{\boldsymbol{\omega}}[\langle\boldsymbol{\omega},\mathbf{x}\rangle\langle\boldsymbol{\omega},\mathbf{z}\rangle] = \mathbb{E}_{\boldsymbol{\omega}}\left[\sum_{i,j=1}^{3n} \omega_i\omega_j \mathbf{x}_i\mathbf{z}_j \right] \nonumber \\
= \sum_{i,j=1}^{3n} \mathbb{E}_{\boldsymbol{\omega}} [\omega_i\omega_j] \mathbf{x}_i\mathbf{z}_j = \sum_{i,j=1}^{3n}\delta_{ij} \mathbf{x}_i\mathbf{z}_j = \langle \mathbf{x}, \mathbf{z} \rangle, \nonumber
\vspace{\cm}
\end{align}
where $\delta_{ij}$ denotes the Kronecker symbol.
\endproof

Once sampled a random number $N \in \mathbb{N}$ with probability $\dfrac{1}{p^{N+1}},$ define $\boldsymbol{\Psi}(\mathbf{x}) = \frac{1}{\sqrt{M}}[\Psi_1(\mathbf{x}),\dots,\Psi_M(\mathbf{x})]$ where $\Psi_1,\dots,\Psi_M$ are all identical copies of the function
\vspace{\cm}
\begin{equation}
\label{eq:zizzi}
\mathbf{x} \longmapsto \sqrt{a_N p^{N+1}} \prod_{j=1}^N \langle \boldsymbol{\omega}_j,\mathbf{x}\rangle,
\vspace{\cm}
\end{equation}
where $\boldsymbol{\omega}_1,\dots,\boldsymbol{\omega}_N$ are independently distributed according to $\boldsymbol{\omega}.$ Equation \eqref{eq:zizzi} and Lemma \ref{lm:lm} allow to extend to our case \cite[Lemma 7]{kk}, which states that the linear kernel $\langle \boldsymbol{\Psi}(\mathbf{x}), \boldsymbol{\Psi}(\mathbf{z}) \rangle$ obtained through  $\boldsymbol{\Psi}$ is an unbiased estimator of the original function $k(\mathbf{x},\mathbf{z}).$ Similarly, using the same arguments of Section 4.1 in \cite{kk}, we obtain that $\langle \boldsymbol{\Psi}(\mathbf{x}), \boldsymbol{\Psi}(\mathbf{z}) \rangle \approx k(\mathbf{x},\mathbf{z})$ uniformly over any compact set of $\mathbb{R}^{3n}$. 

Since we proved that $\boldsymbol{\Psi}$ approximates the kernel $k$ in the sense explained above, the final stage is solving the issue related to \eqref{eq:assumption}.

\vspace{\cm}
\begin{prop}
	\label{prop:ppp}
	The map $\boldsymbol{\Psi}$ satisfies the assumption \eqref{eq:assumption}, that is, for every i = $1,\dots,M$, it results
	\vspace{\cm}
	\begin{align}
	\label{eq:xxx}
	\frac{1}{\sqrt{M}}[\Psi_1(\mathbf{h}_i),\dots,\Psi_M(\mathbf{h}_i)] = \mathbf{e}_i.
	\vspace{\cm}
	\end{align}
\end{prop}

\proof
The relationship \eqref{eq:xxx} displays a system of equations, stochastically dependent on the randomness of $\boldsymbol{\Psi}$. Actually, in our case, it is enough to solve the system \eqref{eq:xxx} and prove the existence of $\mathbf{h}_1,\dots,\mathbf{h}_M$ under a specific realization of $N$ and $\boldsymbol{\omega}$, the two sources of randomness in $\boldsymbol{\Psi}$. In other words, we can solve \eqref{eq:xxx} in a maximum likelihood sense by considering the samples of $N$ and $\boldsymbol{\omega}$ which verify \eqref{eq:xxx} with probability $1$. Thus, we use a prior on $N$ so that $N=1$ and, once absorbed into $\mathbf{h}_i$ all the multiplicative constant defining $\boldsymbol{\Psi}$, then \eqref{eq:xxx} becomes
\vspace{\cm}
\begin{equation}
\label{eq:foffo}
[\langle \boldsymbol{\omega}_1, \mathbf{h}_i \rangle,\dots,\langle \boldsymbol{\omega}_M, \mathbf{h}_i \rangle] = \mathbf{e}_i, \quad i = 1,\dots,M.
\vspace{\cm}
\end{equation}
Precisely, \eqref{eq:foffo} is a linear system of size $M$ in the $M$ unknowns $\mathbf{h}_i$. If we then assume that the Dirac delta distribution of  $\boldsymbol{\omega}_j$ is concentrated in $j$ with probability 1, \eqref{eq:foffo} is solvable if and only if $\langle \boldsymbol{\omega}_j, \mathbf{h}_i\rangle = \delta_{ij}$ for any $i,j = 1,\dots,M$. This is actually verified once chosen $\mathbf{h}_i$ to be the $i$-th element of the orthonormal basis of $\mathbb{R}^{3n}$.
\endproof

With Proposition \ref{prop:ppp}, all issues related to the computability for $\widehat{\mathbb{S}}(k)$ is solved. Additionally, one can also easily understand that, with the previous choice of $\mathbf{h}_i$, once selected a linear kernel $k(\mathbf{x},\mathbf{z}) = \langle \mathbf{x},\mathbf{z}\rangle$, then $\widehat{\mathbb{S}}(k)$ is equal to the $\widehat{\mathbb{S}}(\mathbf{X})$, so that the classical covariance is a particular case of our framework.

The theoretical discussion leads to derive Algorithm 1 and to apply the proposed kernelized covariance for the task of action and activity recognition.  For a better understanding, we also visualize such pipeline in Figure \ref{fig:pipeline}.

\vspace{.2 cm}
\noindent {\bf Computational cost.} The complexity of our trial-specific kernelized covariance is $O(M^2T^2)$. Thus, differently from previous approaches \cite{MsB,Jayasumana:CVPR2013,Minh,Harandi:CVPR14}, the proposed framework is very efficient if compared to the cubic complexity of methods like \cite{Jayasumana:CVPR2013} which require eigen-decomposition. Under a mathematical point of view, our kernelized covariance is a natural generalization of the classical covariance matrix, which can be retrieved as a particular case in our paradigm once fixed the kernel function \eqref{eq:serieker} to be a linear one. On the other hand, the computational cost still remains the same if compared with the classical covariance descriptor.

\begin{figure}[t!]
	\vspace{\cm}
	\vspace{\cm}
	\vspace{\cm}
	\vspace{\cm}
	\includegraphics[width=\columnwidth,keepaspectratio]{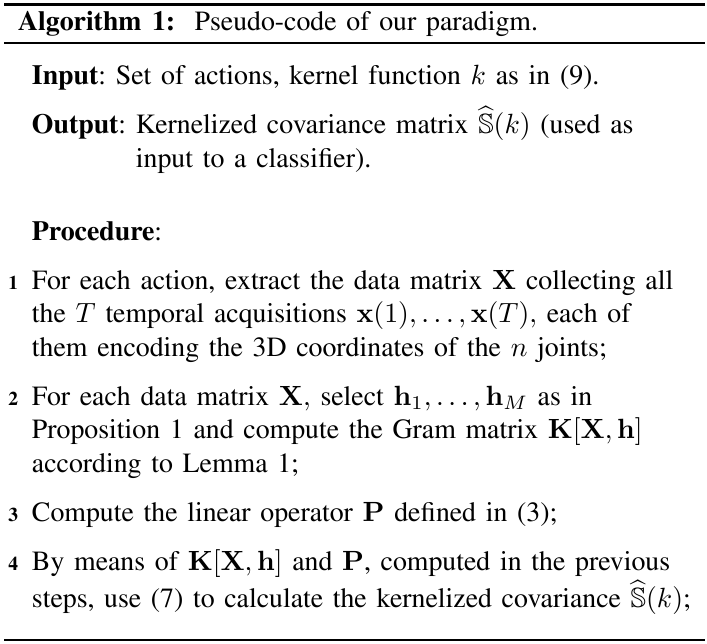}
	\label{alg1}
\end{figure}

\begin{table*}[t!]
	\vspace{\cm}
	\centering
	\caption{Comparative performance of the proposed kernelized-COV benchmarking previous methods in the literature \cite{TPM:ECCV06,egizi,Harandi:CVPR14} based on covariance matrix and the state-of-the-art approach \cite{beyond}. Best results in bold.}
	\begin{tabular}{r|c|c|c|c|}
		Method & {MSR-Action3D} & {MSR-Daily-Activity} & {MSRC-Kinect12} & {HDM-05}\\\hline\hline
		Region-COV \cite{TPM:ECCV06} & 74.0\% & 85.0\% & 89.2\% & 91.5\%  \\
		Hierarchy of COVs \cite{egizi} & 90.5\% & - & 91.7\% & - \\
		COV-$J_\mathcal{H}$-SVM \cite{Harandi:CVPR14} & 80.4\% & 75.5\% & 89.2\% & 82.5\%  \\
		Ker-RP-POL \cite{beyond} & 96.2\% & {\bf 96.9\%} & 90.5\% & 93.6\%  \\
		Ker-RP-RBF \cite{beyond} & {\bf 96.9\%} & 96.3\% & 92.3\% & 96.8\%  \\\hline
		\textbf{Kernelized-COV} (proposed) & 96.2\% & 96.3\% & {\bf 95.0\%} & {\bf 98.1\%} \\
	\end{tabular}
	\label{tab:vsall}
	\vspace{1 cm}
\end{table*}

\vspace{\cm}
\vspace{\cm}
\section{Experimental results}\label{sez:exp}

In this section, we present the experimental results obtained with our \textit{Kernelized-COV} method on different publicly available MoCap datasets for action recognition. Precisely, the following algorithms were compared in our experiments:  \textit{Region-COV} \cite{TPM:ECCV06} (covariance region descriptor), temporal pyramid of covariance descriptors (\textit{Hierarchy of COVs}) \cite{egizi} and, finally, an infinite covariance operator which exploits Bregman divergence, namely \textit{COV-$J_\mathcal{H}$-SVM} \cite{Harandi:CVPR14}. Furthermore, we also report the comparison against the recent state-of-the-art methods, namely \textit{Ker-RP-POL} and \textit{Ker-RP-RBF} \cite{beyond}. 

In all the experiments, we followed \cite{beyond} in performing SVM classification by means of a global log-Euclidean kernel applied upon Gram matrices, directly computed over joints coordinates, encoding each single trial. Nevertheless, differently from \cite{beyond}, in order to represent each multivariate time series of joints trajectories, the data encoding of any trial was realized through our kernelized covariance matrix $\widehat{\mathbb{S}}(k)$, where $k$ is the exponential-dot product kernel (see Section \ref{sez:method}). For a fair comparison, our kernelization was plugged into the publicly available code\footnote{\url{http://www.uow.edu.au/~leiw/}} and, for classification, we used the \textit{SVM and Kernel Methods Matlab Toolbox}\footnote{\url{http://asi.insa-rouen.fr/enseignants/~arakoto/toolbox/index.html}} using the wrapper directly provided by the authors. Finally, we fixed $M=3n$ and, as done by \cite{beyond}, the kernel parameter $\sigma > 0$ is chosen by cross validation.


In all the experiments, we only used the 3D skeleton coordinates available in the following datasets:

\begin{itemize}
	\item {MSR-Action3D} \cite{Action3D}, where there are 20 classes of mostly sport-related action (\emph{e.g.}, \emph{jogging} or \emph{tennis-serve}) involving 10 subjects. Since each subject performs each action 2 or 3 times, the  overall number of trials is 567. For each of them, Kinect sensor is used to acquire depth maps, from which 20 joints are extracted to model the human pose of any of the human agents. 
	\item {MSR-Daily-Activity} \cite{Daily}, captured by using a Kinect device and it is composed by 16 different classes related to every-day actions such as  \emph{read book} or \emph{lie down on sofa}. All of them are performed by 10 subjects. The main difficulty of this dataset originates from the fact that any activity class is performed in an either standing/sitting position, with a consequent misleading motion pattern to mess up the classification.
	\item {MSRC-Kinect12} \cite{MSRC}, consisting of sequences of human movements, represented as body-part locations, and the associated gesture to be recognized by the system. 594 sequences of approximate total length of six hours and 40 minutes are collected from 30 people performing 12 gestures: in total, 6,244 gesture instances. The motion files contain Kinect estimated trajectories of 20 joints. 
	\item {HDM-05} \cite{HDM-05}, containing more than tree hours of systematically recorded and well-documented MoCap data using a 240Hz VICON system to acquire the gestures of 5 non-professional actors via 31 markers. Motion clips have been manually cut out and annotated into roughly 100 different motion classes: on average, 10-50 realizations per class are available.
\end{itemize}

In all cases, we used the same splits adopted in \cite{beyond}: for MSR-Action3D, MSR-Daily-Activity and MSRC-Kinect12, training is performed on odd-index subject, while the even-index ones are left for testing (cross-subject pipeline of \cite{Action3D}), while, in HDM-05, the training split exploits all the data from the ``\texttt{bd}'' and ``\texttt{mm}'' subjects and testing is performed on ``\texttt{bk}'', ``\texttt{dg}'' and ``\texttt{tr}''.

Furthermore, for the HDM-05 dataset we removed some severely corrupted samples \cite{egizi} and, as performed by \cite{beyond}, selected only the following classes:  \emph{clap above head}, \emph{deposit floor}, \emph{elbow to knee}, \emph{grab high}, \emph{hop both legs}, \emph{jog}, \emph{kick forward}, \emph{lie down floor}, \emph{rotate both arms backward}, \emph{sit down chair}, \emph{sneak}, \emph{squat}, \emph{stand up lie} and \emph{throw basketball}. 
All the data are pre-processed in a common way. In particular, in MSR-Action3D and MSR-Daily-Activity, we computed the velocity and acceleration from the raw positions of the joints adopting either first and second order finite different scheme respectively as in \cite{movpose}. 

Table \ref{tab:vsall} shows the results of \textit{Kernelized-COV} on the four different datasets in comparison with all the other methods. Therein, in the case of MSR-Action3D and MSR-Daily-Activity, our proposed method is able to achieve comparable results with a small deviation from the state-of-the-art \cite{beyond}, but it outperforms all the other competitors.
More impressively, on MSRC-Kinect12, \textit{Kernelized-COV} improves the-state-of-the-art \cite{beyond} by $2.7\%$. Even in the last dataset, namely HDM-05, the accuracy of the proposed method is $1.3\%$ higher of the best score achieved by the other competitors. In this case, referring to \cite{egizi}, we did not report the accuracy on HDM-05 due to the different experimental settings: \textit{Hierarchy of COVs} scored $95.41\%$ on a simplified $11$-class problem, while, in the same conditions, we scored $98.8\%$. Furthermore, it is worth noting that, on all the considered datasets our \textit{Kernelized-COV} works even better than a recent infinite covariance operator \cite{Harandi:CVPR14}, more discriminatively encoding the data.

The improvements in classification accuracies demonstrate the effectiveness of \textit{Kernelized-COV}. Moreover, our proposed principled way of encoding non-linearities conveyed by the data is always superior to classical covariance based methods such as \cite{TPM:ECCV06,egizi,Harandi:CVPR14} and does not suffer the gap in performance showed by covariance representation in \cite{beyond}. 

\begin{table}[h!]
	\vspace{\cm}
	\centering
	\caption{Comparison against other classical approaches for action and activity recognition from MoCap data.}
	\label{tab:altri}
	\begin{tabular}{r|c|}
		Method & {MSR-Action3D} \\ \hline\hline
		Action Graph \cite{Li:CVPRw10} & 79.0\% \\
		Random Occupancy Patterns \cite{Wang:ECCV12} & 86.0\% \\
		Actionlets \cite{Wang:CVPR12} & 88.2\% \\
		Pose Set \cite{pose} & 90.0\% \\
		Moving Pose \cite{movpose} & 91.7\% \\
		Lie Group \cite{Vemulapalli:CVPR14} & 92.5\% \\
		Normal Vectors \cite{SNV} & 93.1\% \\ \hline
		\textbf{Kernelized-COV} (proposed) & {\bf 96.2\%} 
	\end{tabular}
	\vspace{.7 cm}
\end{table} 


As a final remark, it is interesting to compare the performance of our \textit{Kernelized-COV} with other not covariance-based methods. To this aim, we take into account the MSR-Action3D dataset and we compared with many previous approaches in the literature, already introduced in Section \ref{sez:intro}. From this analysis, the results presented in Table \ref{tab:altri} give a further evidence of the effectiveness of the proposed use of the kernelized covariance, which is able to overcome \cite{SNV}, the best score reported, by a margin of 3.1\%.

\vspace{\cm}
\vspace{\cm}
\vspace{\cm}
\section{Conclusions \& Future Perspectives}\label{sez:conc}

This paper presents a principled mathematical paradigm to recover the applicability of kernel trick for covariance matrix, in order to better model more general class of relationships other than the linear ones. This enhances the descriptiveness of the classical covariance matrix which is retrievable as a particular case of our general theoretical framework. Experimentally, \textit{Kernelized-COV} closes the gap between covariance and kernel-based representations in many action recognition datasets, namely MSR-Action3D, MSR-Daily-Activity, MSRC-Kinect12 and HDM-05. The proposed method is able to improve the previous best accuracies, setting the new state-of-the-art performance on the last two datasets.

As a future work, we either tackle the applicability of this novel framework to other classification problems and we will also investigate how a similar pipeline can be extended to more general classes of kernel functions.

\vspace{\cm}\vspace{\cm}

\balance

\bibliographystyle{IEEEtran}
\bibliography{fonti}

\end{document}